%% file: main.tex
\documentclass[10pt, logo, twocolumn, copyright, nonumbering]{nvidiatechreport}

\usepackage[utf8]{inputenc} 
\usepackage[T1]{fontenc}    
\usepackage{hyperref}       
\usepackage{url}            
\usepackage{booktabs}       
\usepackage{amsfonts}       
\usepackage{nicefrac}       
\usepackage{microtype}      
\usepackage[dvipsnames]{xcolor}         
\usepackage{multirow}
\usepackage{multicol}
\usepackage{graphicx}
\usepackage[numbers]{natbib}
\usepackage{xspace}
\usepackage{adjustbox}
\usepackage{tcolorbox}
\usepackage{enumitem}
\usepackage{wrapfig}
\usepackage{subcaption}    
\usepackage{float}
\usepackage{stfloats}
\usepackage{amsmath}
\usepackage{listings}
\usepackage{mdframed}
\usepackage{arydshln}
\usepackage{footnote}
\tcbuselibrary{listings,breakable}

\definecolor{NvidiaGreen}{RGB}{118, 185, 0}

\usepackage[utf8]{inputenc} 
\usepackage[T1]{fontenc}    
\usepackage{hyperref}       
\usepackage{url}            
\usepackage{booktabs}       
\usepackage{amsfonts}       
\usepackage{nicefrac}       
\usepackage{microtype}      
\usepackage{xcolor}         
\usepackage{multirow}
\usepackage{graphicx}

\newcommand{\dataset}{\texttt{Nemotron-Math}}

\title{Nemotron-Math: Efficient Long-Context Distillation of Mathematical Reasoning from Multi-Mode Supervision} 

\author{
\centering {Wei Du, Shubham Toshniwal, Branislav Kisacanin, Sadegh Mahdavi, Ivan Moshkov, George Armstrong, Stephen Ge, Edgar Minasyan, Feng Chen, Igor Gitman
}
}

\begin{abstract}
\textbf{Abstract:} 
High-quality mathematical reasoning supervision requires diverse reasoning styles, long-form traces, and effective tool integration, capabilities that existing datasets provide only in limited form. Leveraging the multi-mode generation ability of \texttt{gpt-oss-120b}, we introduce \dataset{}, a large-scale mathematical reasoning dataset containing 7.5M solution traces across high, medium, and low reasoning modes, each available both with and without Python tool-integrated reasoning (TIR). The dataset integrates 85K curated AoPS problems with 262K community-sourced StackExchange-Math problems, combining structured competition tasks with diverse real-world mathematical queries. We conduct controlled evaluations to assess the dataset’s quality. \dataset{} consistently outperforms the original \texttt{OpenMathReasoning} on matched AoPS problems, and incorporating StackExchange-Math substantially improves robustness and generalization, especially on HLE-Math, while preserving accuracy on math-competition benchmarks. To support efficient long-context training, we develop a sequential bucketed strategy that accelerates 128K context-length fine-tuning by 2–3× without significant accuracy loss compared to the full-length training. To verify the scalability of our supervision, we further perform experiments on \texttt{Qwen3-8B} and \texttt{Qwen3-30B-A3B}, showing that both models converge to similar final accuracy under our full context training recipe. Overall, \dataset{} provides diverse, high-quality, and scalable reasoning supervision, enabling state-of-the-art performance, including 100\% \texttt{maj@16} accuracy on AIME 2024/2025 for both \texttt{Qwen3-8B} and \texttt{Qwen3-30B-A3B} with Python TIR.

\end{abstract}

\begin{document}

\maketitle

\input{tex/1_introduction}

\input{tex/2_overview}

\input{tex/3_experiment}

\input{tex/4_results}

\input{tex/5_related_work}

\section{Conclusion}
We introduce \dataset{}, a large-scale mathematical reasoning dataset containing 7.5M long-form solution traces generated by \texttt{gpt-oss-120b} across multi-mode and tool-augmented settings, covering 347K curated problems. By combining structured AoPS problems with diverse StackExchange-Math questions, \dataset{} provides broad domain coverage and rich reasoning variability. Experiments show that it offers higher-quality supervision than prior datasets and improves both competition-level and open-domain reasoning. To enable efficient long-context training, we proposed a sequential bucketed strategy that delivers 2–3× faster 128K-context fine-tuning with only 1–3\% accuracy difference from full joint training. Models fine-tuned on \dataset{} also scale effectively: both \texttt{Qwen3-8B} and \texttt{Qwen3-30B-A3B} achieve 100\% \texttt{maj@16} accuracy on AIME24/25 under the Python TIR high reasoning mode. We will release all data, code, and trained models to support reproducibility and further open-source development. 

\bibliographystyle{plainnat}
\bibliography{paper}

\clearpage
\appendix

\input{tex/6_appendix}

\end{document}

%% file: tex/1_introduction.tex
\section{Introduction}

Mathematical problem-solving remains a key benchmark for evaluating the reasoning capabilities of large language models (LLMs).
Unlike general natural language processing tasks, mathematical problems often require multi-step logical deduction, symbolic manipulation, and long-context understanding.
Recent large-scale datasets, such as OpenMathInstruct-2 \cite{toshniwal2024openmathinstruct2}, Skywork-MathQA \cite{zeng2024skyworkmathdatascalinglaws}, and NuminaMath \cite{li2024numinamath}, have substantially advanced the study of mathematical reasoning in LLMs. Building on these efforts, subsequent datasets such as BackMATH \cite{zhang2025backmath} and OpenMathReasoning \cite{moshkov2025aimo} extend this progress toward competition-level and olympiad-style reasoning.

Despite these advances, most existing mathematical reasoning datasets are generated by single mode reasoning models, which produce relatively uniform solution styles and limited variation in reasoning depth or tool usage. At the same time, many recent efforts have focused primarily on increasing the difficulty of competition-style mathematical problems. While highly effective for constructing challenging reasoning data, these datasets tend to focus on formal, competition-style problems that cover a relatively narrow range of mathematical domains. As a result, current mathematical reasoning datasets capture correctness and complexity but only partially reflect the broader spectrum of reasoning behaviors encountered across diverse mathematical problems. 

The recently released GPT-OSS family of open-weight reasoning models, with \texttt{gpt-oss-120b}~\cite{agarwal2025gpt} as the flagship, offers a new opportunity to address these limitations. Unlike prior models, it provides three controllable reasoning modes, high, medium, and low, that produce solutions of varying depth and length, and it can generate exceptionally detailed tool-integrated reasoning traces through extensive Python calls. These capabilities make it possible to construct datasets that capture diverse reasoning styles, self-verification behaviors, and tool-usage patterns that were previously inaccessible.

Building on this opportunity, we construct \dataset{}\footnote{\url{https://huggingface.co/datasets/nvidia/Nemotron-Math-v2}}, a large-scale mathematical reasoning dataset designed to combine rich reasoning diversity with long-context supervision. On the data side, we curate a high-quality mathematical problem set by combining structured, competition-style questions from \texttt{OpenMathReasoning}~\cite{moshkov2025aimo} with diverse, community-driven questions from Math Stack Exchange~\cite{mathse} and MathOverflow~\cite{mathoverflow}, which together constitute the StackExchange-derived portion (StackExchange-Math) of our dataset. Following prior work, we remove proof-style items and further filter out problems that are too easy for \texttt{gpt-oss-120b}: for each question, the model generates 16 solutions with low reasoning mode (8 with and 8 without Python tool-integrated reasoning (TIR)), and any problem with a high pass rate (>= 0.8) is discarded as trivial. This filtering reduces the Art of Problem Solving (AoPS)~\cite{AoPS} portion, corresponding to the AoPS problems included in ~\cite{moshkov2025aimo}, from the original 175K problems to 85K, and reduces the StackExchange-Math  portion from 651K to 262K, resulting in a balanced collection of challenging and nontrivial tasks.

From this filtered problem pool, we retain the multi-mode trajectories generated by \texttt{gpt-oss-120b} under high, medium, and low reasoning modes, each both with and without Python TIR. After removing trajectories that fail to reach the reference answer, the final \dataset{} corpus contains 7.5M high-quality long-form reasoning traces up to 128K tokens in length, capturing diverse reasoning depths, self-verification styles, and tool-usage behaviors. These properties make \dataset{} a rich resource for studying long-context and tool-augmented mathematical reasoning.

To further enhance the efficiency of long-context fine-tuning on such large-scale data, we propose a sequential bucketed training strategy that groups samples by sequence length and trains the model progressively from 16K to 128K tokens.
This staged approach enables optimized parallelism configurations at each length scale, significantly improving training throughput and resource utilization.
Although it may introduce minor accuracy trade-offs compared to full-length joint training, the method achieves 2–3× faster training while maintaining strong overall performance.

Building on these developments, our main contributions are summarized as follows:
\begin{itemize}
    \item We present \dataset{}, a large-scale mathematical-reasoning dataset that contains 7.5M long-form solution traces produced by \texttt{gpt-oss-120b} under three distinct reasoning modes (high, medium, and low), both with and without Python TIR. By integrating 85K curated AoPS problems with 262K diverse StackExchange-Math questions, the dataset provides broad domain coverage and rich variability in reasoning depth, style, and tool usage.

    \item Through controlled comparisons, we show that \dataset{} provides substantially higher-quality supervision than prior datasets. Under the high reasoning mode setting without Python TIR, it improves AIME25 \texttt{pass@1} for \texttt{Qwen3-30B-A3B} by 13.1\% over the baseline. Incorporating StackExchange-Math further enhances robustness, especially on HLE-Math, without sacrificing competition performance.

    \item We propose a sequential bucketed training strategy that progressively expands model context from short to long (128K) windows. This approach improves training throughput by 2–3× while maintaining accuracy within 1–3\% of full-length joint training, making ultra-long-context fine-tuning computationally practical.

    \item We conduct scaling studies on \texttt{Qwen3-8B} and \texttt{Qwen3-30B-A3B}, showing that both architectures benefit from \dataset{} and converge to similar final performance. Under the high reasoning mode with Python TIR, both models achieve 100\% \texttt{maj@16} accuracy on AIME24 and AIME25, demonstrating the strength and generality of the dataset.
\end{itemize}

%% file: tex/2_overview.tex
\section{Nemotron-Math Overview}
The \dataset{} dataset integrates diverse mathematical problems from multiple sources, combining problems from the \texttt{OpenMathReasoning} dataset~\cite{moshkov2025aimo}, which is based on the AoPS community, with additional problems collected from the Stack Exchange-math \cite{mathse,mathoverflow}.
Each problem is used as a prompt to the \texttt{gpt-oss-120b} model, which generates multiple solution trajectories under different reasoning modes, including both with and
without Python TIR. 
In total, this process yields approximately 7.5M reasoning traces. 
Detailed descriptions of the data sources, generation pipeline, quality filtering, and dataset statistics are provided in the following subsections.

\subsection{Problem Set}

The problem set of \dataset{} is constructed from two complementary sources, designed to balance curated mathematical rigor with real-world diversity.

\subsubsection{AoPS Source}
The first source is adopted from the \textit{OpenMathReasoning} dataset~\cite{moshkov2025aimo}, which was constructed from problems collected on the AoPS community forum~\cite{AoPS}. For convenience, we refer to this subset as the AoPS Source throughout the paper. 
To ensure that the tasks in our dataset admit verifiable final answers, we exclude items whose primary objective is theorem proving and retain only nontrivial, high-quality problems with checkable solutions. The resulting subset contains approximately 175K challenging mathematical questions spanning algebra, geometry, number theory, and combinatorics, and is further reduced to 85K by removing easy questions as described in Section~\ref{sec:solution_generation}.

\subsubsection{StackExchange-Math Source}
The second component, StackExchange-Math Source\footnote{
We use only the StackExchange data dumps released before the July 2024 policy change,
when the content was distributed under CC BY-SA without additional usage restrictions.},
is constructed from Math Stack Exchange~\cite{mathse} and MathOverflow~\cite{mathoverflow}.
These platforms, similar to AoPS in being user-generated and partially overlapping in audience, tend to feature more college-level and research-oriented problems.
We apply the same preprocessing and filtering procedures as in \citet{moshkov2025aimo}. 
Proof-style questions are filtered following the procedure described in \citet{moshkov2025aimo}, which employs the \texttt{Qwen2.5-32B-Instruct} model as a classifier. 
In particular, the same decontamination process is applied to eliminate any overlap with public benchmarks, ensuring consistency with the previous setup.
After all filtering and validation steps, the StackExchange-Math Source yields approximately 651K distinct mathematical problems, which are further reduced to 262K after removing easy questions as described in Section~\ref{sec:solution_generation}.

\begin{table*}[t]
\centering
\begin{tabular}{llcc}
\toprule
\multicolumn{2}{c}{} & \multicolumn{2}{c}{Data Size (in K)} \\
\cmidrule(lr){3-4}
Tool Usage & Reasoning Mode & AoPS & StackExchange-Math \\
\midrule
\multirow{3}{*}{Python TIR}
 & High   & 466 & 1106 \\
 & Medium & 401 & \phantom{1}945 \\
 & Low    & 271 & \phantom{1}719 \\
\midrule
\multirow{3}{*}{Without Python TIR}
 & High   & 438 & 1086 \\
 & Medium & 354 & \phantom{1}899 \\
 & Low    & 191 & \phantom{1}621 \\
\bottomrule
\end{tabular}
\caption{Distribution of the 7.5M generated solutions across tool usage and reasoning modes for the AoPS and StackExchange-Math problem sets (in K).
The AoPS set includes 85K problems, and the StackExchange-Math set includes 262K problems.}
\label{tab:data_stats_joint}
\end{table*}

\subsection{Solutions Generation}
\label{sec:solution_generation}
Based on the prepared problem set, we prompt the \texttt{gpt-oss-120b} model to generate solutions in three reasoning modes, high, medium, and low, under both Python TIR and without Python TIR settings, yielding six configurations in total. For each configuration, eight solutions are generated by varying random seeds with a temperature of 1.0 and top-p of 1.0. To ensure correctness, we follow \citet{moshkov2025aimo} and use \texttt{Qwen2.5-32B-Instruct} \cite{hui2024qwen2} as an LLM-as-a-judge, comparing the expected answers with the final answers extracted from the generated solutions. All solution-generation and evaluation steps are orchestrated through the \texttt{Nemo-Skills} framework \cite{nemo-skills}.

For each problem, we use model-generated solutions to verify the extracted answers from the source forums and to replace them when necessary, ensuring reliable reference answers. Specifically, we generate 16 high reasoning mode solutions (8 Python TIR and 8 without Python TIR) using \texttt{gpt-oss-120b}, as this mode provides the highest data quality. When a problem has no extracted answer from the forum, we use the majority vote among these model solutions. When an extracted answer is available, we retain it as long as at least one model solution is judged consistent with it; if all model solutions disagree, we replace the extracted answer with the majority vote of the model-generated answers. Manual inspection of the cases where replacement occurs indicates that the extracted answers are typically noisy or incomplete, whereas the majority-vote \texttt{gpt-oss-120b} solutions are more reliable.

To further filter out trivial problems, we computed the pass rate of each problem using low-reasoning solutions (8 Python TIR and 8 without Python TIR attempts). Problems with a pass rate above 0.8 were removed, as such items are typically too easy for the model and contribute limited training signal. 
After applying this filtering, the AoPS portion was reduced from 175K to 85K problems, and the StackExchange-Math portion from 651K to 262K. Finally, we discard any generated solutions that fail to reach the expected answer, ensuring that the resulting dataset emphasizes challenging and high-quality problem–solution pairs.

\begin{table*}[t]
\centering
\begin{tabular}{lcccc}
\toprule
Data Source & $\leq$16K & 16K–32K & 32K–64K & $\geq$64K \\ 
\midrule
AoPS & \phantom{1}175K & 235K & 114K & 23K \\
StackExchange-Math & 5022K & 254K & \phantom{1}82K & 20K \\
\bottomrule
\end{tabular}
\caption{Distribution of the 7.5M generated solutions by reasoning trace length bucket across AoPS and StackExchange-Math sources. 
Bucket ranges (e.g., 16K–32K) denote token lengths, and values indicate the number of reasoning traces (in K).}
\label{tab:bucket_stats}
\end{table*}

\subsection{Solutions Analysis}
The statistics of the 7.5M generated solutions is summarized in Table~\ref{tab:data_stats_joint}. The Python TIR variants are consistently larger than the variants without Python TIR, as code execution during reasoning enables the model to solve more problems. Moreover, the overall data size increases from the low reasoning to the high reasoning mode, reflecting the fact that more reasoning effort can aid solution quality. These patterns arise directly from our final filtering step, in which we remove any solutions that do not lead to the expected answer. The data distribution statistics by bucket length are presented in Table \ref{tab:bucket_stats}, which indicates that the dataset is heavily skewed towards shorter sequences.

%% file: tex/3_experiment.tex
\section{Experiments Setup}
\subsection{Training Details}
We fine-tune \texttt{Qwen3-8B} and \texttt{Qwen3-30B-A3B}  \cite{qwen3technicalreport} using a consistent supervised training pipeline across all experiments. The model is optimized with AdamW \cite{DBLP:conf/iclr/LoshchilovH19} with a fixed learning rate of 2e-4, selected based on a learning-rate sweep (Appendix \ref{sec:lr_sweep}), without warmup, and a global batch size of 2048. Sequence packing is applied to improve efficiency on long-context reasoning data.  All components of the pipeline, including problem extraction, data generation, training, and evaluation, are orchestrated using \texttt{Nemo-Skills} \cite{nemo-skills}. During training, \texttt{Nemo-Skills} leverages \texttt{NeMo-RL} \cite{nemo-rl} with the Megatron backend.

\subsection{Evaluation Setup}
We evaluate our models on two complementary benchmark suites. 
(1) Comp-Math-24-25~\cite{moshkov2025aimo}, which includes the HMMT-24-25, AIME24, and AIME25 subsets, 
covers competition-style mathematical reasoning problems emphasizing symbolic precision and multi-step deduction. 
(2) the text-only Math subset of the recently released HLE (Humanity’s Last Exam) benchmark~\cite{Phan2025HLE}, a multi-domain evaluation suite featuring both text and multimodal problems.
The HLE-Math subset consists of 976 diverse problems designed to assess high-level mathematical reasoning. It spans algebra, geometry, combinatorics, and calculus, and often demands long-context comprehension and verification of intermediate steps. Together, Comp-Math-24-25 and HLE-Math form a balanced evaluation protocol that covers both formal competition-style reasoning and open-domain advanced mathematical problem solving.

For each model, we evaluate the final fine-tuned checkpoint under six configurations: high, medium, and low reasoning modes, each in both with Python TIR and without Python TIR settings. For the benchmarks AIME24, AIME25, HMMT-24-25, we generate 16 solutions per problem using temperature 1.0, top-p 1.0, and a maximum generation length of 120K tokens. For HLE-Math, we use the same decoding setup but generate 4 solutions per problem, reflecting its much larger number of questions (976 in total) and empirically low cross-seed variance (< 1\%).  We report both \texttt{pass@1} and \texttt{maj@k}, where \texttt{pass@1} measures the average accuracy across different independent runs and \texttt{maj@k} computes accuracy under majority voting over $k$ generated solutions
(with $k=16$ for AIME24, AIME25, HMMT-24-25 and $k=4$ for HLE-Math).

For AIME24, AIME25, and HMMT-24-25, we use math-verify for automatic numeric/symbolic answer checking. For HLE-Math, we instead rely on an LLM-as-a-judge protocol (\texttt{Qwen2.5-32B-Instruct}) using our evaluation prompt (Appendix \ref{sec:LLM_prompts}). For comparison, we also evaluate the baseline \texttt{Qwen3-8B} and \texttt{Qwen3-30B-A3B} models using their default decoding configuration~\cite{qwen3technicalreport} (temperature 0.6, top-p 0.95, maximum generation length 120K tokens).

%% file: tex/4_results.tex
\section{Experimental Results}
To evaluate the effectiveness of our dataset construction, we compare \dataset{} with the updated version of \texttt{OpenMathReasoning}~\cite{moshkov2025aimo}. 
The original OpenMathReasoning dataset was initially generated using \texttt{DeepSeek-R1}, and its generation pipeline was later updated to \texttt{DeepSeek-R1-05-28}\footnote{\url{https://huggingface.co/datasets/nvidia/Nemotron-Post-Training-Dataset-v1/viewer/default/math}}. 
Our comparison is conducted against this updated version. 
In contrast, \dataset{} regenerates all reasoning trajectories using \texttt{gpt-oss-120b} and additionally incorporates problems from StackExchange-Math, providing broader coverage beyond AoPS.

To ensure a fair and controlled comparison, we construct two aligned datasets from the same pool of 50K unique AoPS problems shared across both datasets. 
We evaluate \dataset{} using its high-reasoning, without Python TIR subset, so that both datasets exhibit comparable reasoning depth and sequence length.
For each selected problem, we collect all corresponding reasoning records from both datasets and retain an equal number of examples by matching the smaller count on each side. This yields two balanced datasets, each containing 264K examples. We also construct a mixed dataset by randomly sampling half of the examples from each source, again totaling 264K examples, to examine possible complementarity between the two reasoning styles. This controlled setup isolates the effect of dataset design and reasoning behavior within the without Python TIR setting while keeping data scale and problem distribution identical.

\begin{table*}[t]
    \centering
    \begin{tabular}{lccc}
    \toprule
    Configuration & AIME24 & AIME25 & HMMT-24-25 \\
    \midrule
    \dataset{} & 81.04 (90.00) & 77.08 (90.00) & 63.17 (73.43) \\
\texttt{OpenMathReasoning} \cite{moshkov2025aimo} &  71.04 (82.50) & 59.38 (71.67) & 49.30 (63.21) \\
     \texttt{Mixed} &  76.46 (88.33) & 66.25 (86.67) & 55.71 (66.70)  \\
    \bottomrule
    \end{tabular}
    \caption{Accuracy of \texttt{Qwen3-30B-A3B} fine-tuned with different datasets on three benchmarks (without Python TIR, high reasoning mode). Metric: \texttt{pass@1} (\texttt{maj@16}, \%).}
    \label{tab:openmath_vs_openmath2}
\end{table*}

We evaluate the final models fine-tuned from \texttt{Qwen3-30B-A3B} without Python TIR, high reasoning mode setting,  as both subset datasets used in this comparison contain only data of this type.  Table~\ref{tab:openmath_vs_openmath2} reports the final evaluation results, measured by \texttt{pass@1} and \texttt{maj@16} on the HMMT-24-25, AIME24, and AIME25 benchmarks. As shown in the table, models trained on \dataset{} consistently outperform those trained on the original \texttt{OpenMathReasoning} and on the \texttt{Mixed} dataset across all benchmarks.  The \texttt{Mixed} dataset, which combines equal portions of \dataset{} and \texttt{OpenMathReasoning}, also surpasses the previous dataset \texttt{OpenMathReasoning},  indicating that the reasoning traces in \dataset{} enable models to exhibit stronger reasoning ability and more effective problem-solving behavior.

\subsection{Effect of StackExchange-Math Integration}
To isolate the contribution of community-sourced problems, we construct two
controlled subsets \emph{within} \dataset{}, which we refer to as \textbf{AoPS-only} and \textbf{AoPS+StackExchange-Math}. AoPS-only contains all AoPS problems and their corresponding solution traces. AoPS+StackExchange-Math is obtained by randomly replacing half of the AoPS examples (problem–solution pairs) with examples drawn from StackExchange-Math, while keeping the total number of examples fixed. In this way, AoPS+StackExchange-Math differs from AoPS-only primarily through the inclusion of StackExchange-Math-based solutions, allowing us to assess the impact of community-sourced content.

StackExchange-Math introduces broader linguistic variation, more informal phrasing, and richer real-world mathematical reasoning patterns compared to the structured, competition-style AoPS problems. This controlled setup therefore enables a clean evaluation of whether incorporating community-driven supervision improves reasoning robustness and out-of-distribution generalization, particularly on open-domain benchmarks such as HLE-Math, without compromising accuracy on competition-style tasks.

\begin{table*}[t]
    \centering
    \resizebox{0.95\textwidth}{!}{
    \begin{tabular}{llllll}
    \toprule
    Dataset & Configuration 
        & AIME24 & AIME25 & HMMT-24-25 & HLE-Math \\
    \midrule
    \multirow{6}{*}{\textbf{AoPS-only}} 
        & High, w/o Python TIR        & 86.88 (95.00) & 83.96 (90.00) & 70.06 (77.99) & 12.22 (12.22) \\
        & High, Python TIR   & 92.67 (100.00) & 94.89 (100.00) & 81.53 (86.19) & 22.54 (22.54) \\
        & Medium,  w/o Python TIR      & 61.88 (75.00) & 71.04 (81.67) & 52.04 (64.01) & 7.22 (7.22) \\
        & Medium, Python TIR & 79.11 (86.67) & 84.89 (100.00) & 72.72 (80.39) & 13.40 (13.40) \\
        & Low,  w/o Python TIR         & 51.25 (63.33) & 41.25 (54.44) & 31.44 (39.48) & 3.92 (3.92) \\
        & Low, Python TIR    & 66.00 (76.67) & 63.11 (76.67) & 58.16 (67.94) & 7.91 (7.91) \\
    \midrule
    \multirow{6}{*}{\parbox{2.8cm}{\textbf{AoPS +\\ StackExchange-Math}}}

        & High,  w/o Python TIR        & 90.00 (94.17) & 86.40 (96.67) & 71.40 (77.85) & 13.60 (13.60) \\
        & High, Python TIR   & 94.58 (100.00) & 97.29 (100.00) & 81.76 (85.28) & 24.67 (24.67) \\
        & Medium,  w/o Python TIR      & 71.88 (83.33) & 67.71 (80.83) & 55.84 (66.22) & 8.58 (8.58)\\
        & Medium, Python TIR & 83.12 (88.33) & 84.17 (100.00) & 73.41 (77.55) & 15.22 (15.22) \\
        & Low,  w/o Python TIR         & 52.50 (71.67) & 45.00 (59.44) & 33.64 (43.99) & 4.10 (4.10) \\
        & Low, Python TIR    & 66.67 (80.00) & 63.75 (76.88) & 58.80 (68.02) & 7.99 (7.99) \\
    \bottomrule
    \end{tabular}
    }
    \caption{Accuracy comparison of \textbf{AoPS-only} and \textbf{AoPS+StackExchange-Math} subsets  of \dataset{} under six reasoning configurations  (high, medium, and low reasoning modes, each evaluated with both Python TIR and without Python TIR setting). Results are reported on AIME24, AIME25, HMMT-24-25, and HLE-Math.  For AIME24/25 and HMMT-24-25, metrics are computed using 16 sampled solutions per problem (\texttt{pass@1} (\texttt{maj@16}, \%)). For HLE-Math, which contains substantially more questions (976 in total) and exhibits extremely low cross-seed variance (typically $<1\%$), results are averaged over 4 seeds while maintaining comparable stability (\texttt{pass@1} (\texttt{maj@4}, \%)). 
}

    \label{tab:stackexchange-math_effect}
\end{table*}

Final results are presented in Table~\ref{tab:stackexchange-math_effect}. 
Across all six reasoning configurations 
(high, medium, and low reasoning modes, each evaluated under both Python TIR and without Python TIR setting),
models fine-tuned on the AoPS-only subset of \dataset{} and those trained on the AoPS+StackExchange-Math subset exhibit clear and consistent trends. For competition-style benchmarks (AIME24, AIME25, and HMMT-24-25), the AoPS+StackExchange-Math variant achieves accuracy that is either comparable to or slightly higher than the AoPS-only variant across all modes, indicating that incorporating StackExchange-Math data does not hinder the model’s ability to solve highly structured olympiad-style problems. In contrast, the gains on HLE-Math are consistent across all configurations. Since HLE-Math contains community-driven, open-domain mathematical questions with diverse linguistic expressions and less formalized reasoning structures, its distribution more closely aligns with StackExchange-Math. Training with the AoPS+StackExchange-Math subset therefore provides more suitable supervision for such tasks, leading to improved robustness overall.

\subsection{Sequential Bucketed Training Strategy}
As shown in Table~\ref{tab:bucket_stats}, the majority of reasoning traces in  \dataset{} fall within short or medium context lengths, with relatively few examples exceeding 64K tokens. Training the full model directly with a fixed 128K context window is therefore highly inefficient, 
as it forces all optimization steps to use the most memory-intensive and communication-heavy parallelism settings, 
despite most samples not requiring such long contexts.

To address this mismatch, we adopt a sequential bucketed training strategy. The dataset is partitioned into buckets according to sequence length, 
and training proceeds in stages from short to long contexts 
(16K $\rightarrow$ 32K $\rightarrow$ 64K $\rightarrow$ 128K). 
At each stage, parallelism configurations (tensor, pipeline, and context parallelism; see Appendix~\ref{sec:train_config} for per-bucket settings) 
are tailored to the current maximum sequence length, ensuring efficient memory usage and significantly reduced communication overhead. Sequence packing is applied throughout to maximize GPU utilization. This progressive approach offers two practical advantages.  
First, for the same short-context data, early stages can operate with extremely high throughput by using parallelism configurations optimized for short sequences. For example, when training on the 16K bucket, an optimized configuration runs at approximately 18 seconds per step, whereas forcing the same 16K data to use the parallelism setup required for 128K context increases the step time to around 25 seconds. This comparison isolates the effect of parallelism configuration, showing that most training tokens can be processed under substantially cheaper settings in the early stages. Second, only the final stage must operate with the full 128K context window, 
meaning that relatively few optimization steps require the expensive long-context parallelism setup. 
In practice, this yields a 2–3$\times$ reduction in end-to-end training cost without altering the final target context length.

\begin{table*}[t]
    \centering
    \resizebox{1.0\textwidth}{!}{%
    \begin{tabular}{lcc|cc|cc}
    \toprule
    \multirow{2}{*}{Configuration} 
    & \multicolumn{2}{c|}{AIME24} 
    & \multicolumn{2}{c|}{AIME25} 
    & \multicolumn{2}{c}{HMMT-24-25} \\
     & Bucket Training & Full Data 
     & Bucket Training & Full Data 
     & Bucket Training & Full Data  \\
    \midrule
    High, without Python TIR         & 86.67 (93.33) & 88.12 (97.50) & 84.17 (96.67) & 84.79 (95.00)  & 68.78 (77.78) & 71.43 (79.78) \\
    High, Python TIR   & 93.54 (100.00) & 94.67 (100.00) & 95.83 (100.00) & 96.00 (100.00) & 78.28 (85.68) & 80.82 (85.05) \\
    Medium, without Python TIR      & 72.92 (80.00) & 76.46 (88.33) & 63.96 (78.33)  & 67.71 (76.67) & 52.17 (65.98) & 55.10 (64.54) \\
    Medium, Python TIR  & 83.54 (87.78) & 83.33 (88.33) & 81.04 (96.67)  & 86.67 (100.00) & 70.76 (79.85) & 73.50 (79.16) \\
    Low, without Python TIR         & 52.08 (70.00) & 50.62 (64.00) & 40.42 (58.82)  & 44.38 (61.11) & 32.33 (42.11) & 32.27 (41.49) \\
    Low, Python TIR    & 68.96 (80.00) & 66.46 (76.67) & 60.21 (78.76)  & 64.58 (77.50) & 57.17 (67.30) & 59.06 (66.50) \\
    \midrule
\texttt{Qwen3-30B-A3B} (baseline) & \multicolumn{2}{c|}{81.25 (90.00)} & \multicolumn{2}{c|}{71.67 (80.00)} & \multicolumn{2}{c}{57.33 (64.97)} \\
\bottomrule
    \end{tabular}
    } 
    \caption{
    Accuracy comparison between full-length joint training and sequential bucketed training on three benchmarks 
    (AIME24, AIME25, and HMMT-24-25) using \texttt{Qwen3-30B-A3B}. Results are reported for three reasoning modes (high, medium, low) under both Python TIR and without Python TIR settings. The last row shows the accuracy of the pretrained \texttt{Qwen3-30B-A3B} model before any training.
    }
    \label{tab:30b_bucket_results}
\end{table*}

Final results obtained using \texttt{Qwen3-30B-A3B} are summarized in Table~\ref{tab:30b_bucket_results}. 
Across all benchmarks and reasoning configurations, the sequential bucketed training strategy achieves accuracy that is broadly comparable to full-length joint training. Many configurations match the full-data results almost exactly, while others show only a minor 1–3\% degradation. This demonstrates that the bucketed strategy preserves nearly the same level of accuracy while substantially reducing overall training cost. As expected, deeper reasoning modes consistently outperform shallower ones, and Python TIR models achieve the strongest results across all benchmarks. Under the high reasoning mode, the fine-tuned model substantially improves over the base \texttt{Qwen3-30B-A3B}. For example, on 
AIME25 (without Python TIR), accuracy improves 13.1\% (from 71.67\% to 84.79\%), highlighting the benefit of supervised reasoning traces. Notably, the high reasoning, Python TIR configuration reaches \texttt{maj@16} accuracy of 100\% on both AIME24 and AIME25, demonstrating the combined effect 
of deeper reasoning depth and tool-augmented supervision.

While these results confirm that sequential bucketed training preserves strong performance while delivering significant efficiency gains, we note that this strategy should be applied with care.
In particular, the distribution of reasoning modes across buckets can become highly imbalanced at long context lengths.
Since medium and low reasoning solutions rarely reach 128K tokens, naively training only on high-reasoning samples in the final stage can cause the model’s behavior to collapse toward uniformly long, high-depth reasoning.
In our experiments, such imbalance leads medium and low modes to generate increasingly long sequences and lose their intended distinction, even though overall accuracy may increase.
To mitigate this effect, we explicitly balance the final long-context stage by sampling a small proportion of medium and low reasoning data, ensuring that all reasoning modes remain visible throughout training.
This practice preserves mode diversity while retaining the efficiency and performance benefits of the sequential bucketed strategy.

\subsection{Scaling with Model size and Architecture}
\begin{figure*}[t]
    \centering
    \begin{minipage}[t]{0.48\linewidth}
        \centering
        \includegraphics[width=\linewidth]{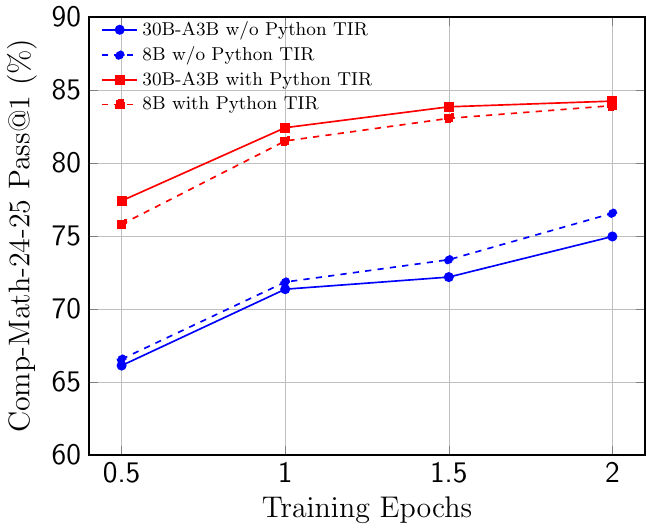}
    \end{minipage}
    \hspace{0.01\linewidth}
    \begin{minipage}[t]{0.48\linewidth}
        \centering
        \includegraphics[width=\linewidth]{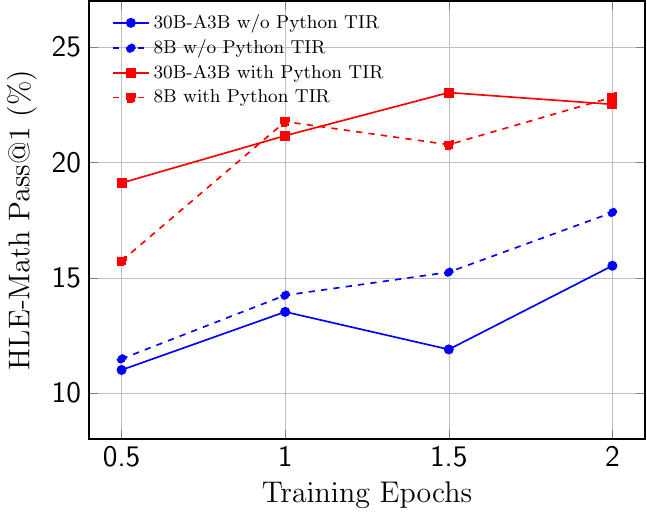}
    \end{minipage}

\caption{
Scaling with model size and architecture on \dataset{}.
Each panel reports \texttt{pass@1} as a function of training progress (evaluated every half epoch),
comparing \texttt{Qwen3-8B} and \texttt{Qwen3-30B-A3B} under the high reasoning mode.
The left panel shows results on Comp-Math-24-25 and the right panel shows results on HLE-Math.
Within each panel, both without Python TIR and with Python TIR settings are plotted.
}

    \label{fig:8B_30B_high}
\end{figure*}

To study how \dataset{} scales across model sizes and architectures,
we compare \texttt{Qwen3-8B} and \texttt{Qwen3-30B-A3B} trained under identical data and training configurations.
Figure~\ref{fig:8B_30B_high} reports results under the high reasoning mode, with evaluations performed every half epoch.

The left panel shows performance on Comp-Math-24-25, while the right panel shows performance on HLE-Math.
In each panel, results without Python TIR and with Python TIR are plotted together.
Solid lines with markers denote \texttt{Qwen3-30B-A3B}, while dashed lines with matching marker shapes denote \texttt{Qwen3-8B}; blue curves correspond to the without Python TIR setting, and red curves correspond to the with Python TIR setting. For clarity, only the high reasoning mode is shown, as the medium and low reasoning modes exhibit the same qualitative trends.

Across both benchmarks and tool settings, the two models exhibit highly similar learning dynamics.
They improve at comparable rates and converge to nearly identical final accuracy.
The only noticeable deviation occurs on HLE-Math without Python TIR, where \texttt{Qwen3-8B} attains slightly higher accuracy.
Overall, these results indicate that \dataset{} provides sufficiently strong supervision for both model scales,
leading to aligned convergence behavior on both competition-style and open-domain mathematical reasoning tasks.

%% file: tex/5_related_work.tex
\section{Related Work}
\subsection{Long Chain-of-Thought Reasoning Data}
With the rapid advancement of open-source reasoning models, a growing body of work focuses on enhancing the reasoning capabilities of LLMs by generating long chain-of-thought (CoT) traces, especially for mathematical problem solving. Early datasets such as OpenMathInstruct-1~\cite{toshniwal2024openmathinstruct}, OpenMathInstruct-2~\cite{toshniwal2024openmathinstruct2}, Skywork-MathQA~\cite{zeng2024skyworkmathdatascalinglaws}, and NuminaMath~\cite{li2024numinamath} were among the first to scale mathematical CoT supervision. While these datasets significantly improved mathematical reasoning performance, their solution traces remain relatively shallow, offering limited multi-step deduction, self-correction, or verification.

Recent advances in reasoning models, exemplified by DeepSeek-R1~\cite{guo2025deepseek} and the Qwen3 series~\cite{yang2025qwen3}, have made it possible to produce substantially longer, more coherent, and more systematically structured reasoning trajectories. These models can produce detailed multi-step derivations, engage in extensive self-reflection and consistency checking, leading to more reliable multi-step derivations. Leveraging such capabilities, OpenMathReasoning~\cite{moshkov2025aimo} released 540K olympiad-level problems paired with 3.2M long-form solutions, establishing the state-of-the-art and winning the AIMO-2 competition.

As the field continued to advance, additional datasets emerged to further enrich the landscape of long-form mathematical reasoning. OpenR1-Math-220K~\cite{openr1_math_2025} provides two to four DeepSeek-R1 trajectories per problem across 220K challenging math questions. Zhao et al.~\cite{zhao20251} introduced AM-DeepSeek-R1-Distilled, containing 1.4M question–response pairs with associated thinking traces for general reasoning tasks. DeepMath-103K~\cite{he2025deepmath} offers 103K mathematically verifiable problems, each paired with three distinct DeepSeek-R1 reasoning paths.

\subsection{Tool-Integrated Reasoning}
Tool-Integrated Reasoning (TIR) enhances the problem-solving capability of LLMs, particularly in mathematics, by enabling models to invoke external computational tools, such as Python execution or symbolic solvers, rather than relying solely on natural-language reasoning. This allows LLMs to handle tasks requiring high numeric precision, symbolic manipulation, or multi-step verification more reliably.

A pioneering effort in this direction is Program of Thoughts (PoT)~\cite{chen2022program}, which expresses reasoning steps as executable programs and uses the resulting program outputs to derive final answers, achieving state-of-the-art results across several mathematical reasoning benchmarks. Following this idea, numerous studies have explored how to teach LLMs to use tools effectively.
ToRA~\cite{gou2023tora} integrates natural-language reasoning with external computational libraries and symbolic solvers, achieving significant improvements over open-source baselines.
MathSensei~\cite{das2024mathsensei} extends this direction by combining multiple tools, including web retrieval, Python execution, and symbolic solvers, and reports a 13.5\% accuracy improvement over GPT-3.5-turbo with standard CoT on the MATH dataset.
MathCoder2~\cite{lu2024mathcoder2} introduces a 19.2B-token tool-centric mathematical pretraining corpus that substantially improves downstream reasoning ability when used to train modern base models.
More recently, Torl~\cite{li2025torl} proposed an RL-based framework that teaches base LLMs to autonomously employ computational tools, outperforming prior TIR models by a large margin (17\%).

Despite advances in long-form and tool-integrated reasoning, existing datasets typically rely on single-mode solution generation, offering limited diversity in reasoning depth or tool-usage behavior. \dataset{} instead provides multi-mode supervision (high/medium/low) under both Python-free and Python-augmented settings, producing a broad spectrum of reasoning styles, from concise heuristic chains to deeply structured, tool-driven solutions, yielding a level of behavioral and tool-integration diversity that, to our knowledge, is not present in prior mathematical reasoning datasets.

%% file: tex/6_appendix.tex
\onecolumn
\begin{table*}[t]
    \centering
    \begin{tabular}{l l cccc c}
    \toprule
    Model & Context length & TP & CP & PP & ETP & EMP \\ 
    \midrule
    \multirow{4}{*}{\texttt{Qwen3-30B-A3B}} 
        & 16k  & 4 & 2 & 1 & 1 & 4  \\
        & 32k  & 4 & 4 & 1 & 1 & 8 \\
        & 64k  & 4 & 8 & 1 & 1 & 8 \\
        & 128k & 4 & 8 & 1 & 1 & 8 \\

    \bottomrule
    \end{tabular}
    \caption{Training configuration by bucket length.}
    \label{tab:speed_by_bucket}
\end{table*}

\begin{table*}[t]
    \centering
    \begin{tabular}{l l cccc c}
    \toprule
    Model/Context Length & 16K & 32K & 64K & 128K & Full context 128K & Speedup\\ 
    \midrule

    \texttt{Qwen3-30B-A3B}
        & 117248  & 58988 & 56801 & 22197 & 559802 & 2.2 \\
    \bottomrule
    \end{tabular}
    \caption{Training time breakdown by bucket length and stage, with full-context total and overall speedup.}
    \label{tab:time_by_bucket}
\end{table*}

\begin{table*}[b]
\centering
\resizebox{0.7\textwidth}{!}{
\begin{tabular}{lcccc}
\toprule
\multirow{2}{*}{Benchmark} &
\multicolumn{2}{c}{With Python TIR } &
\multicolumn{2}{c}{Without Python TIR } \\
\cmidrule(lr){2-3} \cmidrule(lr){4-5}
& \texttt{pass@1} & \texttt{maj@k} & \texttt{pass@1} & \texttt{maj@k} \\
\midrule
AIME25 & 95.62\% & 100.00\% & 83.75\% & 93.33\% \\
AIME24 & 94.38\% & 100.00\% & 88.75\% & 98.33\% \\
HMMT24–25 & 80.29\% & 80.29\% & 70.12\% & 78.91\% \\
HLE-math & 10.32\% & 10.32\% & 10.96\% & 10.96\% \\
\bottomrule
\end{tabular}}
\caption{
Performance of the \texttt{Nemotron-3-Nano-30B-A3B} \emph{SFT-only} checkpoint
(i.e., the pre-RL model used in the Nemotron-3 Nano post training pipeline)
on Comp-Math-24-25 and HLE-Math, evaluated under both with and without Python TIR.
This checkpoint is trained using only the high-reasoning-mode portion of \dataset{}.
Metrics report \texttt{pass@1} and \texttt{maj@k}
($k=16$ for Comp-Math-24-25 and $k=4$ for HLE-Math).
}
\label{tab:nemotron}
\end{table*}

\section{Training Configuration By Bucket Length}
\label{sec:train_config}
Table~\ref{tab:speed_by_bucket} summarizes the parallelism settings used for each bucket length in the sequential training schedule.
Because different sequence lengths place pressure on different parts of the training stack, we adapt the parallelism configuration at each stage.
Tensor parallelism (TP) splits the large matrix multiplications inside each layer across devices, while context parallelism (CP) shards the sequence dimension to enable efficient long-context training. Pipeline parallelism (PP) is used only lightly, as most speedup comes from TP and CP in our experiment.
For the MoE model \texttt{Qwen3-30B-A3B}, we additionally employ expert tensor parallelism (ETP) to parallelize computation within each expert, and expert model parallelism (EMP) to distribute different experts across devices.

Short-context buckets (e.g., 16K) can be trained very efficiently when using
parallelism configurations optimized for short sequences.
For the \texttt{Qwen3-30B-A3B} model, a 16K bucket runs at around 18 seconds per step
under its own optimized TP/CP setup.
In contrast, training the same data under the fixed parallelism configuration
required for 128K context length slows execution to approximately 25 seconds per step. We report a detailed breakdown of step time by bucket length and parallelism configuration in Table~\ref{tab:time_by_bucket}.
By running most training steps under parallelism settings tailored to shorter contexts and reserving heavy configurations only for the final 128K stage, the sequential bucketed strategy achieves a 2–3$\times$ end-to-end speedup compared to training the entire dataset with a fixed 128K context window.

\section{Further Evidence via NVIDIA-Nemotron-3-Nano-30B-A3B-BF16 Evaluation}
\label{sec:nemotron_results}

Table~\ref{tab:nemotron} reports results for a \texttt{NVIDIA-Nemotron-3-Nano-30B-A3B-BF16}
\emph{SFT-only} checkpoint\footnote{The final public release of \texttt{NVIDIA-Nemotron-3-Nano-30B-A3B-BF16} (\url{https://huggingface.co/nvidia/NVIDIA-Nemotron-3-Nano-30B-A3B-BF16}) includes additional reinforcement learning stages; the results reported here are based on the pre-RL SFT checkpoint.} from the Nemotron-3 Nano post-training pipeline, before any reinforcement learning stages. This model is supervised fine-tuned using only the high-reasoning-mode trajectories from \dataset{}, and we evaluate it on both Comp-Math-24-25 and HLE-Math under
with and without Python TIR settings. On Comp-Math-24-25, the SFT-only Nemotron checkpoint achieves accuracy comparable to our fine-tuned \texttt{Qwen3-30B-A3B}, suggesting that the high–reasoning-mode subset already provides strong competition-style supervision.
On HLE-Math, \texttt{Qwen3-30B-A3B} performs better, which we attribute to our use of the full \dataset{} including large-scale StackExchange-Math supervision that better matches the open-domain and linguistically diverse nature of HLE-Math.

\section{Learning Rate Grid Search}
\label{sec:lr_sweep}
We performed a grid search for hyperparameter tuning on \texttt{Qwen3-30B-A3B} using a subset of the training data, with the results summarized in Table~\ref{tab:lr_sweep}. 
Overall, a learning rate of 2e-4 generally performs best across different reasoning modes, and we fix this learning rate for all experiments.

\begin{table*}[t]
\centering
\resizebox{1.0\textwidth}{!}{%
\begin{tabular}{lcccccc}
\toprule
Configuration 
& lr=$1e$-$5$ 
& lr=$5e$-$5$ 
& lr=$7e$-$5$ 
& lr=$1e$-$4$ 
& lr=$2e$-$4$ 
& lr=$5e$-$4$ \\
\midrule
High, no tool
& 38.75\% $\pm$ 1.96\%
& 50.46\% $\pm$ 1.80\%
& 50.46\% $\pm$ 1.80\%
& 53.98\% $\pm$ 2.44\%
& 54.39\% $\pm$ 2.28\%
& 48.22\% $\pm$ 1.71\% \\

High, with tool
& 50.85\% $\pm$ 2.20\%
& 65.50\% $\pm$ 2.64\%
& 65.50\% $\pm$ 2.64\%
& 66.20\% $\pm$ 2.00\%
& 68.21\% $\pm$ 2.84\%
& 67.07\% $\pm$ 1.17\% \\

Medium, no tool
& 26.15\% $\pm$ 1.74\%
& 36.18\% $\pm$ 2.46\%
& 36.18\% $\pm$ 2.46\%
& 43.95\% $\pm$ 2.00\%
& 42.75\% $\pm$ 2.22\%
& 32.64\% $\pm$ 1.97\% \\

Medium, with tool
& 49.34\% $\pm$ 1.90\%
& 61.69\% $\pm$ 1.97\%
& 61.69\% $\pm$ 1.97\%
& 66.43\% $\pm$ 1.76\%
& 68.68\% $\pm$ 1.71\%
& 60.52\% $\pm$ 2.51\% \\

Low, no tool
& 17.29\% $\pm$ 1.28\%
& 22.00\% $\pm$ 2.32\%
& 22.00\% $\pm$ 2.32\%
& 27.54\% $\pm$ 1.44\%
& 25.24\% $\pm$ 1.49\%
& 18.31\% $\pm$ 2.30\% \\

Low, with tool
& 40.26\% $\pm$ 2.71\%
& 48.14\% $\pm$ 2.35\%
& 48.14\% $\pm$ 2.35\%
& 53.25\% $\pm$ 1.95\%
& 53.76\% $\pm$ 1.96\%
& 46.34\% $\pm$ 2.10\% \\
\bottomrule
\end{tabular}
}
\caption{Accuracy comparison on the Comp-Math-24-25 benchmark under different learning rates using \texttt{Qwen3-30B-A3B}. Results are reported for three reasoning modes (high, medium, low) under both Python TIR and without Python TIR settings. All results are reported as pass@1 accuracy (mean $\pm$ standard deviation) over 16 runs.}
\label{tab:lr_sweep}
\end{table*}

\section{Answer Judgment Prompt}
\label{sec:LLM_prompts}

\begin{tcolorbox}[breakable,width=\textwidth,colback=white,colframe=NvidiaGreen,title={\centering \large  \textbf{Prompt: Answer Judgment}}]
\footnotesize                  
\begin{lstlisting}[
breaklines=true, postbreak={},breakindent=0pt, 
label={lst:math-prompt}]
user: |-
    You will be asked to look at the two answers (predicted and expected) to a math problem and to judge whether they are equivalent within the context of the problem.

    Please first explain your reasoning in a couple of sentences. Then respond with only Yes or No as your judgement on whether the two answers are the same.
    When comparing answers only perform trivial simplifications.

    Here are a few examples.

    Example 1:
    Problem: Factor $7x^3 - 21x^2 + 14x$.
    Predicted answer: $7x(x - 2)(x - 1)$
    Expected answer: $7x(x-1)(x-2)$

    Reasoning: The order of the factors does not matter, so the answers are the same.
    Judgement: Yes

    Example 2:
    Problem: A rectangle has a length of 6 meters and a width of 2 meters. If the length is reduced by 3 meters and the width is halved, what is the new area of the rectangle in square meters?
    Predicted answer: 3/2
    Expected answer: 1.5

    Reasoning: 3/2 is the same as 1.5
    Judgement: Yes

    Example 3:
    Problem: Simplify the expression $\sqrt{{7!}}$, where $n!$ stands for $n\cdot(n-1)\cdot(n-2)\cdots \cdot 2\cdot 1$.
    Predicted answer: 71
    Expected answer: 12\sqrt{{35}}.

    Reasoning: This is non-trivial to simplify, so the answers are different.
    Judgement: No

    Example 4:
    Problem: What is the simplified form of the expression $\sqrt{{98 x^{{3}} y^{{5}} z}} ?
    \begin{{align*}}
    \text{{A)}} & 2 x y z \sqrt{{7 x y z}} &
    \text{{B)}} &  7 x^{{2}} y^{{2}} \sqrt{{2 y z}}
    \\
    \text{{C)}} & 7 x y^{{2}} \sqrt{{2 x y z}}  &
    \text{{D)}} &49 x y^{{2}} \sqrt{{2 x y z}}
    \\
    \end{{align*}}
    Predicted answer: 7 x y^{{2}} \\sqrt{{2 x y z}}
    Expected answer: C

    Reasoning: Predicted answer is the same as the expected answer choice C.
    Judgement: Yes

    Example 5:
    Problem: A line segment of length $5$ has one endpoint at $(1, 2)$ and the other endpoint at $(4, b)$. Find all possible values of $b$, separated by commas.
    Predicted answer:  -2, 6
    Expected answer: 6, -2

    Reasoning: The order doesn't matter in the context of the problem.
    Judgement: Yes

    Example 6:
    Problem: Solve $\tan x = \sin x$ for $0 \le x \le 2 \pi.$  Enter all the solutions, separated by commas.
    Predicted answer: 0, \pi
    Expected answer: 0,\pi,2\pi.

    Reasoning: Number of solutions is different.
    Judgement: No

    YOUR TASK

    Problem: {problem}
    Predicted answer: {predicted_answer}
    Expected answer: {expected_answer}
\end{lstlisting}
\end{tcolorbox}